\documentclass[runningheads]{llncs}
\pdfoutput=1
\usepackage{eccv}
\usepackage{eccvabbrv}

\usepackage{graphicx}
\usepackage{booktabs}

\usepackage[accsupp]{axessibility}  
\usepackage{hyperref}

\usepackage{orcidlink}

\usepackage{colortbl}     
\definecolor{blue}{RGB}{30,144,255}
\newcommand{\tablestyle}[2]{\setlength{\tabcolsep}{#1}\renewcommand{\arraystretch}
{#2}\centering\footnotesize}
\newlength\savewidth\newcommand\shline{\noalign{\global\savewidth\arrayrulewidth
  \global\arrayrulewidth 1pt}\hline\noalign{\global\arrayrulewidth\savewidth}}
\newcommand{\cmark}{\ding{51}}%
\usepackage{pifont}    
\newcommand{\method}{FOMA\xspace}
\newcommand{\cona}{MFA\xspace}

\usepackage{wrapfig} 

\begin{document}

\title{Focus-Consistent Multi-Level Aggregation for
Compositional Zero-Shot Learning} 

\titlerunning{FOMA}

\author{Fengyuan Dai\inst{1} \and
Siteng Huang\inst{1} \and
Min Zhang\inst{1} \and
Biao Gong\inst{2} \and
Donglin Wang\inst{1}
}

\institute{Westlake University \and Alibaba Group
}

\maketitle


\begin{abstract}

To transfer knowledge from seen attribute-object compositions to recognize unseen ones, recent compositional zero-shot learning (CZSL) methods mainly discuss the optimal classification branches to identify the elements, leading to the popularity of employing a three-branch architecture.
However, these methods mix up the underlying relationship among the branches, in the aspect of consistency and diversity.
Specifically, consistently providing the highest-level features for all three branches increases the difficulty in distinguishing classes that are superficially similar.
Furthermore, a single branch may focus on suboptimal regions when spatial messages are not shared between the personalized branches.
Recognizing these issues and endeavoring to address them, we propose a novel method called \textbf{Focus-Consistent Multi-Level Aggregation (\method)}.
Our method incorporates a \textbf{Multi-Level Feature Aggregation (\cona)} module to generate personalized features for each branch based on the image content. 
Additionally, a \textbf{Focus-Consistent Constraint} encourages a consistent focus on the informative regions, thereby implicitly exchanging spatial information between all branches.
Extensive experiments on three benchmark datasets (UT-Zappos, C-GQA, and Clothing16K) demonstrate that our \method outperforms SOTA.
\end{abstract}

\keywords{Compositional Zero-Shot Learning \and Multi-Level Feature Aggregation \and Focus-Consistent Constraint}

\section{Introduction\label{sec:intro}}

Based on the familiarity with the compositional concepts \textit{red glove} and \textit{blue hat}, humans can easily distinguish \textit{red hat} even when first encountering it.
To endow deep neural networks with the same ability to identify novel complex concepts by composing known components (\textit{i.e.}, \textit{primitives}), compositional zero-shot learning (CZSL)~\cite{misra2017red} has emerged as a prevalent research focus. 
CZSL aims to recognize unseen attribute-object \textit{compositions} after observing seen compositions that contain all the primitives.

Prior methods apply different architectures: some existing works~\cite{atzmon2020causal, karthik2022kg} learn individual primitive classifiers to mitigate the domain gap between the seen and unseen compositions; another line of works~\cite{misra2017red, li2020symmetry} utilize the relationship between the attributes and objects, projecting the image and the semantic compositions into a common embedding space and measuring the compatibility score. To combine the strengths of the above methods, recently there has been a trend towards three-branch architectures that simultaneously address both primitives and compositions~\cite{saini2022disentangling, hu2023leveraging, wang2023learning}. During inference, the predictions of all three branches are integrated for a more robust recognition.

\begin{figure}[t]
\begin{center}
\includegraphics[width=1.\textwidth]{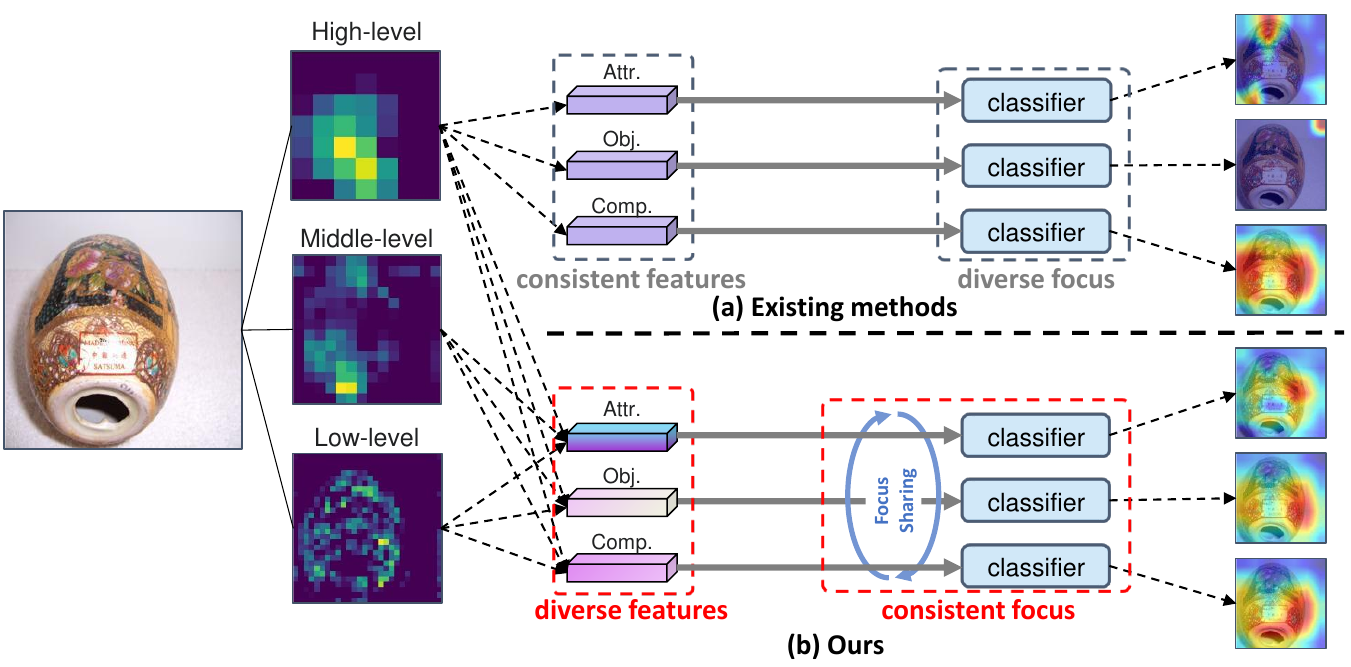}
\end{center}
   \caption{\textbf{Conceptual comparison in terms of consistency and diversity.} Features of the image are inputted into the model, which consists of three classification branches. The heatmap is generated based on the classification process on each branch. On the left, we showcase low-level, middle-level, and high-level features of the image, in which the lower-level features contain more intricate details. On the right, the heatmaps show that constraining branches to focus on the same area mitigates the risk of being misled. }
\label{fig:painted_ceramic}
\end{figure}

Despite the remarkable performance, the underappreciated relationship between the branches in terms of consistency of focus and diversity of features has been mixed up, as depicted in \cref{fig:painted_ceramic}:
\textbf{(1)} All the recognition branches use the consistent highest-level instead of customized visual features, while the global features captured by large receptive fields may not be suitable for accurately classifying certain primitives.
For instance, to recognize an ``\textit{old}'' \textit{man}, low-level texture features can contribute more to detecting wrinkles on a face.
\textbf{(2)} The lack of consistent focus from other branches may leave a single branch confused by the irrelevant region caused by noise or shading. Highly independent and personalized branches often ignore information sharing, particularly common spatial messages that highlight regions beneficial to compositional recognition.


In this paper, we develop a novel model named \textbf{F}ocus-C\textbf{o}nsistent \textbf{M}ulti-level \textbf{A}ggregation (\textbf{\method}) that revisits the relationship between the branches.
Built on the three-branch architecture, \method addresses the aforementioned issues with two non-trivial solutions, i.e. diverse features and consistent focus: 
\textbf{(1)} A \textbf{Multi-Level Feature Aggregation (\cona)} module generates non-uniform features for each branch by aggregating features from different levels.
Since input images can vary in terms of resolution, domain, and recognition target, the weights used to combine multi-level features are instance-specific and determined based on the visual content.
\textbf{(2)} A \textbf{Focus-Consistency Constraint} is applied to establish the correlation among the branches and encourage the joint focus on revealing regions.
Specifically, learnable parameters are introduced to enable the branches to attend to specific areas and
the focusing areas are compared across all the branches through a penalty loss.
This constraint leverages the observation that recognizing attributes and objects can provide additional spatial clues for identifying compositions, and vice versa.

To compare the proposed \method with existing methods, we conduct experiments on three widely used benchmark datasets: UT-Zappos~\cite{yu2014fine}, C-GQA~\cite{naeem2021learning} and Clothing16K~\cite{zhang2022learning}.
Experimental results show that \method, which combines appropriate features and enforces attention on common regions, achieves state-of-the-art performance.
In summary, our contributions are as follows:
\begin{itemize}
\item[$\bullet$] We revisit the previous CZSL works in terms of consistency and diversity in the classification process, and pertinently formulate a novel method named Focus-Consistent Multi-Level Aggregation (FOMA).
\item[$\bullet$] We design a Multi-Level Feature Aggregation module that combines multi-level features for each branch based on the characteristics of visual content, allowing an adaptive and effective recognition.
\item[$\bullet$] We develop a Focus-Consistency Constraint that conveys spatial information across branches by constraining attention to the same regions.
\item[$\bullet$] We conduct extensive experiments on three CZSL datasets, and our \method  outperforms state-of-the-art methods, demonstrating its superiority.
\end{itemize}

\section{Related Work}

\subsection{Compositional Zero-Shot Learning (CZSL)}

The concept of compositionality was first come up by the \cite{hoffman1984parts, biederman1987recognition}. Misra~\etal~\cite{misra2017red} aimed to classify attributes and objects in the context and introduced CZSL problem, which was a branch of zero-shot learning (ZSL)~\cite{lampert2009learning, xian2017zero, chao2016empirical, palatucci2009zero} but recognized compositionality only in the visual domain. Most of the existing works can be roughly divided into two streams.
One stream~\cite{nan2019recognizing, li2020symmetry, ruis2021independent, xu2021zero, naeem2021learning, mancini2021open,yang2020learning, li2022siamese} maps visual and semantic embeddings into the same latent space and calculates their similarity. 
Such combined state-object semantic embeddings can be learned with a transformation function, such as a multi-layer perceptron (MLP)~\cite{misra2017red} or a graph convolutional network~\cite{naeem2021learning}. Another stream~\cite{nagarajan2018attributes, huo2022procc, karthik2022kg, atzmon2020causal, saini2022disentangling} aims to identify the essence of attributes and objects. For example, Nagarajan~\etal~\cite{nagarajan2018attributes} argues that attributes should be learned as operators so that they can express their effects on novel objects efficaciously. OADis~\cite{saini2022disentangling} computes attention between triplet samples and disentangled features according to their similarity and dissimilarity. 

In this work, We emphasise consistency and diversity on the recently popular three-branch structure~\cite{zhang2022learning, Huang:Troika}.
Previous works~\cite{ruis2021independent, xu2021zero, saini2022disentangling} have noticed spatial messages but could not fully explore them. With the help of the Multi-Level Feature Aggregation module and Focus-Consistent Constraint, our \method effectively leverages multi-level features and spatial messages while maintaining the fixed backbone architecture.

\subsection{Visual Attention Consistency}
Visual attention consistency has been extensively explored in various computer vision tasks, such as multi-label image classification~\cite{tsoumakas2007multi, zhang2013review}, domain generalization~\cite{li2017deeper, wang2022generalizing}, and few-shot recognition~\cite{Huang:AGAM}.
Existing methods such as ACfs~\cite{guo2019visual} and ACVC~\cite{cugu2022attention} aim to minimize the differences between attention maps of original and augmented images.
ICASC~\cite{Wang_2019_ICCV} focuses on aligning attention maps from different images of the same class.
Different from these methods, our proposed FOMA calculates the self-attention similarity between the normalization of the composition branch and the other two branches. This encourages different branches to attend to the same regions of the same image.

\subsection{Multi-Level Features}
Different layers of convolutional neural networks (CNNs) extract features at various levels, with lower-level features capturing detailed information and higher-level features conveying semantic meaning~\cite{islam2021shape}. This observation has motivated previous studies~\cite{hosu2019effective, sindagi2019multi, qian2021enhancing, xu2022multi} to emphasize the importance of utilizing multi-level features to enhance performance. In certain detection-focused research~\cite{zhang2017amulet, miao2021pvgnet}, multi-level features have been incorporated into networks by concatenating them in different configurations after feeding them into convolutional layers. However, manually determining the optimal combination of multi-level features is challenging. To address this, we propose training an aggregation predictor to generate sample-oriented features through the fusion of multi-level features.

\section{Methods}

In this section, we begin by presenting the task definition of CZSL.  Secondly, we showcase the overall three-branch framework of \method. Then, a \cona module and Focus-Consistent Constraint are introduced in a detailed manner, elucidating their roles in adequately maintaining consistency and diversity among the branches. Finally, we elaborate on the procedures of inference and training.

\begin{figure*}
\begin{center}
\includegraphics[width=1.\textwidth]{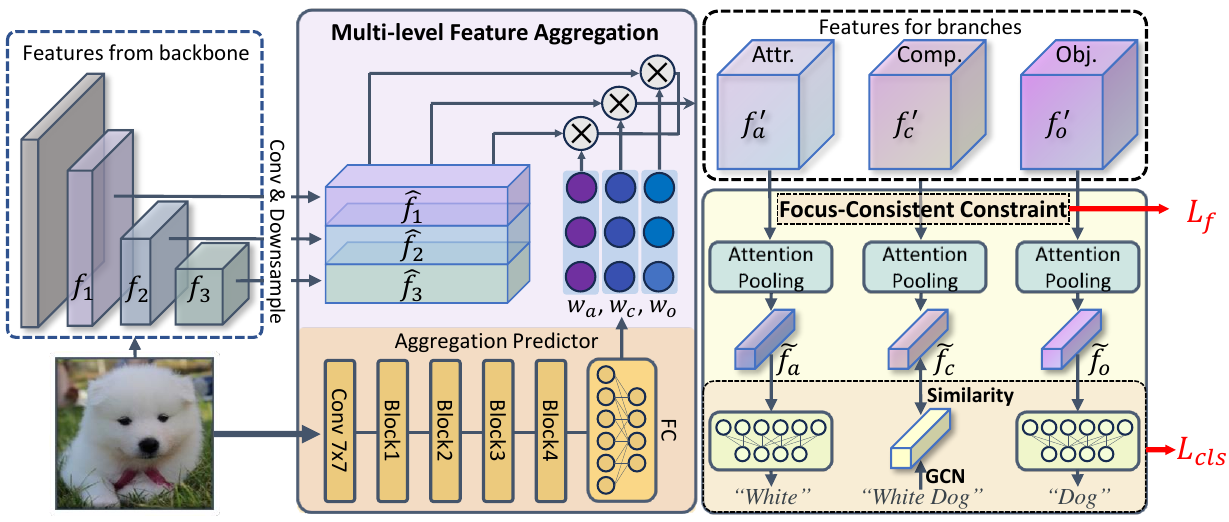}
\end{center}
   \caption{\textbf{The architecture of \method} mainly consists of the three branches, along with \cona module and Focus-Consistent Constraint. 
   Specifically, the image is first inputted into the backbone to extract features. Then MFA module takes both the original image and multi-level features and outputs customized features. 
   After attention pooling, the composition branch computes the similarity between visual and semantic embeddings, while the other branches utilize the classifiers to predict classification labels.
   Focus-Consistent Constraint supervises the classification process across the three branches, which is illustrated minutely in \cref{fig:FCC}. 
       }
\label{fig:framework}
\end{figure*}

\subsection{Task Definition}
We follow the generalized CZSL setup~\cite{purushwalkam2019task}. Denoting input image space as $X$ and label space as $Y$, the dataset $S=\{(x, y)|x\in X, y\in Y \}$ can be defined by the set of tuples of the input image $x$ and label $y$. We split $S$ into two disjoint sets: seen set $S_{s}$ and unseen set $S_{u}$, where $S = S_{s} \cup S_{u}$. Each label $y=(y_a, y_o)$ consists of an attribute $y_a \in A$ and an object $y_o \in O$. It should be noted that the training set only contains a subset of images from $S_{s}$, while validation and testing samples are drawn from $S_{s} \cup S_{u}$.

\subsection{Recognizing in Three Branches \label{sec:3.2}}

We extract features from the backbone and denote them as $f_k \in \mathbb{R}^{C_k\times H_k\times W_k}$, where $k \in {1, 2, 3}$. These correspond to low-level, middle-level, and high-level features\footnote {In this paper, we refer to features that come from the second block, third block, and fourth block of ResNet~\cite{he2016deep} as low-level, middle-level, and high-level feature.}, respectively. 
These three features, along with the original image $x$, are input into the Multi-Level Feature Aggregation (MFA) module, denoted as $MFA(\cdot)$. This process yields features for each branch:
\begin{equation}
f^{\prime}= [f_{a}^{\prime}, f_{c}^{\prime}, f_{o}^{\prime}] = MFA(f_1, f_2, f_3|x),
\end{equation}
where [,] denotes the concatenation.


In the attribute and the object branch, we utilize attention pooling $AP(\cdot)$ to encode spatial messages. Subsequently, a multilayer perceptron with batch normalization is employed as classifier $CLS(\cdot)$ to generate the primitive score:
\begin{equation} 
S_{i} = CLS_i(AP_i(f_{i}^{\prime})), i \in \{a, o\}.
\end{equation}


In the composition branch, we follow CGE~\cite{naeem2021learning} to learn semantic embeddings, projecting semantic and visual embeddings into the same latent space. To be more specific,  we construct an undirected graph with unweighted edges, consisting of $N = |A|+|O|+|Y|$ nodes. Nodes are connected if they belong to the same label set $y = (y_a, y_o), y \in Y$. This allows us to obtain an adjacency matrix $A\in \mathbb{R}^{N\times N}$ and a node embedding matrix $H^{(0)}$. The graph convolution network (GCN) propagates as:
\begin{equation}
H^{(l+1)} = \sigma(D^{-1}\hat{A}H^{(l)}W^{l}),
\end{equation}
where $\sigma$ is the ReLU function and $\hat{A} = A + I_N$ indicating that each node connects itself. $I_N$ is an identity matrix. $W^{l}$ is a learnable matrix and $D$ is a diagonal matrix that is used for normalization. 
The output of  GCN is $H^{(L)}$ and $H_y^{(L)}$ denotes the composition embeddings for compositional classification. 
We take the similarity between visual and semantic embeddings as the composition score:
\begin{equation}
S_{c} = sim(AP(f_{c}^{\prime}), H_y^{(L)}),
\end{equation}
where $sim(\cdot)$ is a similarity function, and it can be implemented in multiple ways. In this case, we employ the dot product for the sake of simplicity.

\subsection{Multi-Level Feature Aggregation}


As pointed out by \cite{matthew2014visualizing, yosinski2015understanding}, due to the characteristics of the receptive field in CNNs, lower layers carry features relating to fine-grained clues like \textit{stripe}, whereas higher-level features are more connected with the information of object classes. This inspires us that some kinds of attributes, e.g. \textit{old} and \textit{wet}, should be more relevant to lower-level features. 
%
Therefore,
we designed a \cona module to aggregate the proper features for the model to better recognize the primitives and compositions. Specifically, \cona module implements an aggregation predictor in CNNs, which takes the original image as input and outputs aggregation weight $w$, indicating multi-features combinations for the attribute, object, and composition branches. We can define the process of \cona as:
\begin{equation}
w =[w_a, w_c, w_o] = softmax(P(x) / \tau ) \in \mathbb{R}^{N_b\times N_f}, \label{for:softmax}
\end{equation}
where $N_b$ and $N_f$ refer to the number of features and branches, with both values set to 3. $P(\cdot)$ denotes the network of aggregation predictor. Additionally, $\tau$ refers to the temperature used in the softmax layer and $w_a$, $w_c$, $w_o$ stand for weights for the attribute, composition and object branch.

For the features extracted directly from the backbone, we first project them into a common space. To be specific, we downsample the spatial dimension of $f_1$ and $f_2$ to $H_3 \times W_3$. Subsequently, we employ a convolution  layer to align the channel dimension.
Noting the convolution layer and the downsample operation with $conv_i(\cdot)$ and $DS(\cdot)$, the aligned feature $\hat{f_k}$ can be defined as:
\begin{equation}
\hat{f_k} = conv_k(DS(f_k)), k \in \{1, 2, 3\}.
\end{equation}

Through stacking and resizing, we can obtain $\hat{f} \in  \mathbb{R}^{N_f\times (C_3\times H_3 \times W_3)}$ from $\hat{f_k}$. Then we use matrix multiplication to aggregate multi-level features:
\begin{equation}
f^{\prime} = w\hat{f}.
\end{equation}

\subsection{Focus-Consistent Constraint}

\begin{wrapfigure}{r}{0.55\linewidth}
\centering
\includegraphics[width=.48\textwidth]{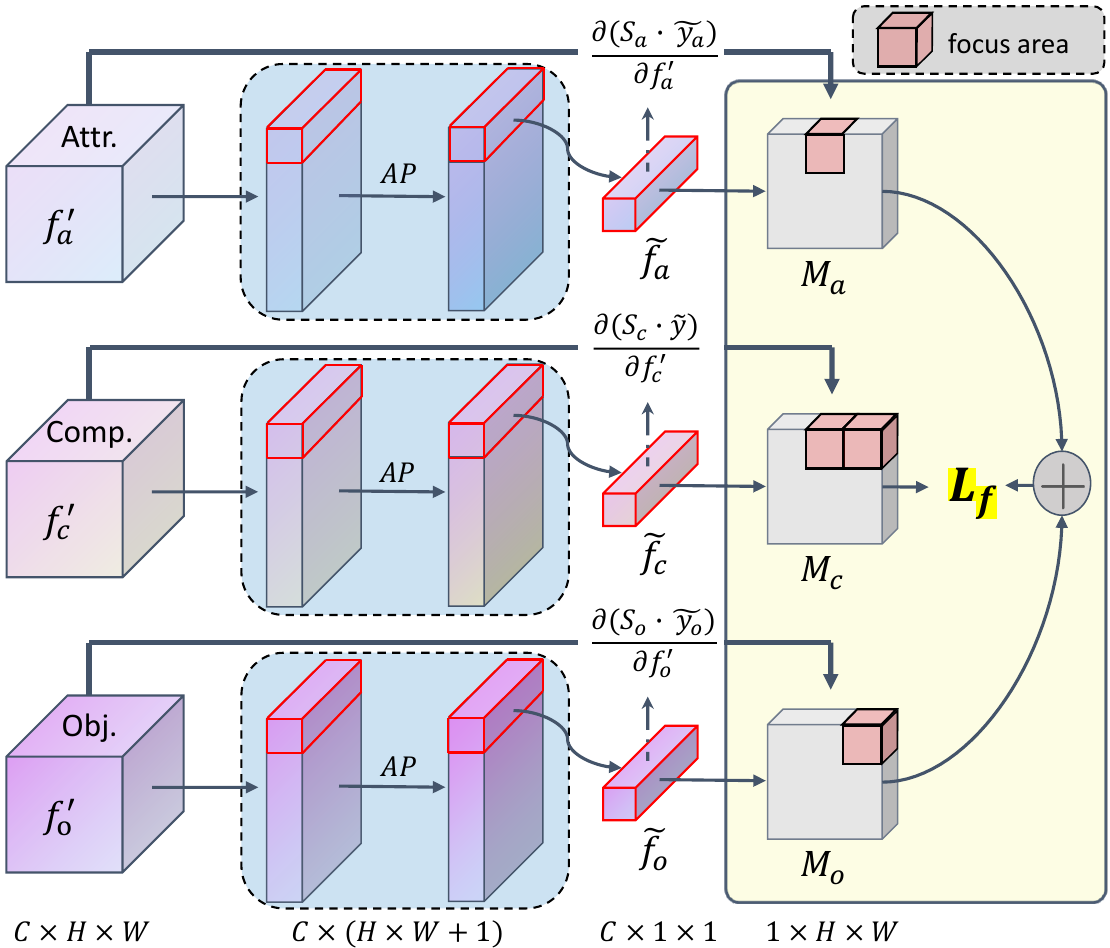}
    \caption{\textbf{Focus-Consistent Constraint.} Features from \cona reduce their spatial dimensions by attention pooling (AP). 
    Visual embeddings $\widetilde{f}$ play a role in determining the score $S$ for each branch. Assigning attention map $M$ as the gradient of $S$ with respect to $f^\prime$ in each branch, we expect the focused region from the composition branch to be equal to the sum of the focused region from the other two branches. 
}
\label{fig:FCC}
\end{wrapfigure}

Like what is discussed before, hidden spatial messages between branches are crucial for guiding the three branches. If one branch is misled by the irrelevant region, these implicitly shared messages could correct the classification process. Previous works do not realize that and merely use global average pooling (GAP) on the output feature of the backbone. GAP is an efficient method to reduce dimensions, however, it inevitably loses vital information. Therefore, we discard the GAP and apply Focus-Consistent Constraint instead to maximize the preservation of revealing messages that are hidden within the spatial dimension.

As shown in \cref{fig:FCC}, given a feature $f \in \mathbb{R}^{C\times H \times W}$, we flatten the spatial dimension and concatenate it with a token, which is initialized to the mean of each channel $f_{GAP}$. Like what is done in ViT~\cite{dosovitskiy2020image}, we add 1-D position embeddings for each patch,  put them together into a self-attention layer and only fetch the first element of the output. During this process, every patch can interact with each other, which fully exploits the information in the spatial domain. We can define attention pooling as:
\begin{equation}
\widetilde{f}= AP(f) = attention([f_{GAP}, f]+PE)[:,0],
\end{equation}
where $\widetilde{f} \in \mathbb{R}^{C}$ denote the output of attention pooling and $PE$ is the 1-D learnable position embeddings.

We follow Grad-CAM~\cite{Selvaraju_2017_ICCV} to gain insight into the specific aspect that each branch focuses on. Specifically, we take the derivative of the score to the corresponding label with respect to $f^\prime$, compress the channel dimension by calculating the mean of each patch and mark it as attention map $M$. Denoting that the $\widetilde{y_c}$, $\widetilde{y_a}$, $\widetilde{y_o}$ are the one-hot vectors of composition $y$, attribute $a$, object $o$, the attention map $M \in \mathbb{R}^{H_3 \times W_3}$ can be defined as:
\begin{gather}
    M_i = mean(\frac{\partial (S_i\cdot \widetilde{y_i})}{\partial f^\prime_i}), i \in \{c, a, o\}.
\end{gather}

We expect the model to pay attention to the informative regions while enabling the attribute branch and object branch to freely choose where they consider important. 
Take \textit{wet dog} for example, humans can infer the attribute \textit{wet} through the part of fur but need to see some distinctive parts on the body of the dog, e.g. \textit{head} or \textit{tail}, before inferring it to be a dog. 
On these grounds, we design the focused loss, which encourages the sum of the focused area from the attribute and the object to be equal to the composition branch:
\begin{equation}
L_{f}= - cos(norm(M_{a}+M_{o}), norm(M_{c})), \label{loss_focus}
\end{equation}
where $\text{norm}(\cdot)$ represents the normalization operation, and $\cos(\cdot)$ denotes the cosine similarity function.

\subsection{Inference and Training}

During the training period, we optimize the parameters of the model with Focus-Consistent Constraint and classification loss. The latter consists of the cross-entropy losses $CE(\cdot)$ from the three branches and can be defined as:
\begin{equation}
\begin{aligned}
 L_{cls} = CE(S_{a}, y_a)+ CE(S_{o}, y_o)+CE(S_{c}, y).
\end{aligned}
\end{equation}

Thus, with the definition of $L_{f}$ in Eq. \eqref{loss_focus}, the overall objective function is formulated as the sum of the two previously mentioned losses:
\begin{equation}
L = L_{cls} + \alpha L_{f}\label{for:total_loss},
\end{equation}
where $\alpha$ is the loss coefficient to balance the two loss terms and is set empirically.

For inference, we first combine multi-level features and obtain branch-specific features through the \cona module. After aggregating spatial information by attention pooling, we utilize two classifiers to recognize the attribute and object. As for composition embeddings, we compute the similarity with semantic embeddings. In the end, we add up the scores from all the branches, which is simple but can effectively reduce the probability of wrong recognition. 

\section{Experiment}


\subsection{Experiment Setup}

\textbf{Datasets. } Previous methods usually use Mit-States~\cite{isola2015discovering} as one of their benchmark datasets. However, as is pointed out by prior works~\cite{zhang2022learning, naeem2021learning}, Mit-States contains a significant amount of noisy labels, making it unqualified as an evaluation dataset. Thus we select UT-Zappos~\cite{yu2014fine}, C-GQA~\cite{naeem2021learning} and Clothing16K\footnote{https://www.kaggle.com/kaiska/apparel-dataset}~\cite{zhang2022learning} for experiments. All of the experiments are done in the closed world setting. 

\begin{table}[t]
\tablestyle{5pt}{1.0}
\setlength\tabcolsep{3.4pt}
\def\w{20pt} 
\scalebox{1.}{
    \begin{tabular}{l|cccc|ccc|ccc}
    &\multicolumn{4}{c}{\textbf{Train}} & \multicolumn{3}{|c|}{\textbf{Validate}} & \multicolumn{3}{c}{\textbf{Test}} \\
    \textbf{Datasets}&Attr. & Obj. & SP & I & SP & UP &I &SP&UP&I  \\
    \shline
    UT-Zappos~\cite{yu2014fine} & 16 & 12 & 83 & 23k & 15 &15 & 3k & 18 & 18 & 3k\\
    C-GQA~\cite{naeem2021learning} & 413 & 674 & 6k & 27k & 1k &1k & 7k & 1k & 1k & 5k\\
    Clothing16K~\cite{zhang2022learning}& 8 & 9 & 18 & 7k & 10 &10 & 5k & 9 & 8 & 3k\\
    \end{tabular}
    }
    \caption{\textbf{The detailed statistics of the datasets.} (SP: \# seen compositions, UP: \# unseen compositions, I: \# images)}
\label{table:datasets_info}
\end{table}

UT-Zappos is a fine-grained dataset with images of shoes. The 12 kinds of shoes have 16 states in total, like \textit{canvas heels} and \textit{leather slippers}. C-GQA is a recently proposed dataset based on the Stanford GQA dataset~\cite{hudson2019gqa}.
Compared to two other datasets, its images have abundant labels with hundreds of attributes and objects. Clothing16K is a dataset for multi-label classification on Kaggle. It consists of 16k images with 8 different clothing categories in 9 different colors. \cref{table:datasets_info} shows the detailed information of the three datasets.

\noindent\textbf{Metrics. }We follow the generalized CZSL setting, aiming to recognize composition whether it is seen or unseen. Prior works~\cite{purushwalkam2019task, chao2016empirical} point out that training on the seen data can cause imbalance when testing on both seen and unseen data. Thus, a calibration bias, which is added directly to the unseen composition results, has been introduced to repair this problem. When the bias varies from $-\infty$ to $+\infty$, we can calculate top-1 seen accuracy (S) and unseen accuracy (U). Besides, we report the value of area under the curve (AUC) and the best harmonic mean (HM). Both of them compute the accuracy of seen and unseen compositions according to the varying bias.

\noindent\textbf{Implementation Details. } Following previous methods, we assign ResNet-18~\cite{he2016deep} pre-trained on ImageNet~\cite{deng2009imagenet} as our backbone and fix it in the training process. We obtain three features from the last three block of ResNet-18 and alter the channel dimension to 512 through the convolution layer with a $1\times1$ kernel. We utilize a lightweight ResNet consisting of two convolution layer in each block to construct the aggregation predictor. 
On all three datasets, we initialize the semantic embeddings with word2vec~\cite{mikolov2013distributed}.

We utilize Adam~\cite{kingma2014adam} optimizer with $5\times 10^{-5}$ learning rate and the same value of weight decay. The learning rate is set using a cosine annealing schedule~\cite{loshchilov2017sgdr}. 
We discuss the introduced hyper-parameters, i.e. the temperature $ \tau$ in Eq.~\eqref{for:softmax} and the loss coefficient $\alpha$ in Eq.~\eqref{for:total_loss}, in \cref{sec:hyper-para}.
\subsection{Main Results}

\begin{table*}[t]
\begin{center}
\begin{tabular}{l|cccc|cccc|cccc} 

 &\multicolumn{4}{c}{\textbf{UT-Zappos}} & \multicolumn{4}{|c|}{\textbf{C-GQA}} & \multicolumn{4}{c}{\textbf{Clothing16K}} \\
\textbf{Methods} & S & U & AUC & HM & S & U & AUC & HM & S & U & AUC & HM   \\
\shline
AttrOpr~\cite{nagarajan2018attributes}&59.8&54.2&25.9&40.8&11.8&3.9&0.3&2.9&93.9&88.3&76.0&77.4 \\
LE+~\cite{misra2017red}&53.0& 61.9& 25.7& 41.0&16.1&5.0&0.6&5.3&93.9&88.3&76.0&77.4 \\
TMN~\cite{purushwalkam2019task} &58.7&60.0&29.3&45.0&21.6&6.3&1.1&7.7&94.9&89.7&79.5&80.9 \\
SymNet~\cite{li2020symmetry} &53.3&57.9&23.9&39.2&25.2&9.2&1.8&9.8&95.7&90.2&73.4&75.2 \\
$\text{CGE}_{\text{ff}}^{\text{\#}}$~\cite{mancini2022learning}&58.8&46.5&21.5&38.0&28.1&10.1&2.3&11.4&94.4&92.9&81.2&80.8 \\
CompCos~\cite{mancini2021open} &59.8&62.5&28.7&43.1&28.1&11.2&2.6&12.4&96.9&93.0&84.7&83.9 \\
IVR~\cite{zhang2022learning}&56.9&65.5&30.6&46.2&-&-&-&-&96.9&94.6&87.0&86.3 \\
$\text{SCEN}^\text{*\#}$~\cite{li2022siamese}&\textbf{63.5}&63.1&32.0&\textbf{47.8}&28.9&12.1&2.9&12.4&97.2&89.9&79.1&78.7 \\ 
SCD~\cite{hu2023leveraging}&62.3&64.5&31.8&46.3&29.9&14.5&3.2&14.1&-&-&-&- \\
CANet~\cite{wang2023learning}&61.0&66.3&\textbf{33.1}&47.3&30.0&13.2&3.3&14.5 &-&-&-&-\\
\shline
\rowcolor[rgb]{.949, .949, .949} \textbf{\method(Ours)}&60.3&\textbf{68.0}&\textbf{33.1}&47.3&\textbf{31.4}&\textbf{15.2}&\textbf{3.8}&\textbf{15.2}&\textbf{97.7}&\textbf{96.0}&\textbf{90.1}&\textbf{88.9} \\
		
\end{tabular}
\end{center}
\caption{\textbf{Comparing \method with state-of-the-art methods:} S, U, AUC, and HM mean best seen compositions accuracy, best unseen compositions accuracy, area under the curve, and harmonic mean of U and S on the test set. The best results are displayed in boldface. The (*) indicates that SCEN on C-GQA is slightly different from other methods. The (\#) means the results on Clothing16K are reported based on the official open-source code.}
\label{table:main_results}
\end{table*}

We compare \method with previous works in \cref{table:main_results}. Except for best seen accuracy and HM on UT-Zappos, our method outperforms all prior works, including AttrOpr~\cite{nagarajan2018attributes}, LE+~\cite{misra2017red}, TMN~\cite{purushwalkam2019task}, SymNet~\cite{li2020symmetry}, $\text{CGE}_{\text{ff}}$~\cite{mancini2022learning},  CompCos~\cite{mancini2021open}, IVR~\cite{zhang2022learning},  SCEN~\cite{li2022siamese}, SCD~\cite{hu2023leveraging} and CANet~\cite{wang2023learning}. SCEN, CANet and IVR are state-of-the-art. For comparability, we follow prior works~\cite{zhang2022learning, li2022siamese} and report the results of all methods with a frozen backbone.

On UT-Zappos, our model achieves the highest result on best unseen accuracy and AUC, indicating that it has a great capacity to balance the attribute and object. However, reminding that UT-Zappos is a fine-grained dataset, the contrastive learning-based method benefits by magnifying the distance between similar labels although it incurs additional time to sample triplets. Therefore, our methods fall behind SCNE~\cite{li2022siamese} on best seen accuracy and HM.

On C-GQA, our model consistently performs better than previous works. Due to the high complexity and large number of compositions on this dataset, there lies a great gap between the best seen and unseen accuracy, indicating the difficulty of recognition on this dataset. But we still outperform state of the art by promoting 0.5 and 0.7 on AUC and~HM. 

The same conclusion applies to the Clothing16K dataset. This dataset, characterized by a limited variety of clothing types and colors, allows all methods to achieve relatively high performance. Still, our method shows a noteworthy improvement across all four metrics. This demonstrates our approach's superiority in generalizing from seen to unseen pairs, even in simpler scenarios.

\subsection{Ablation Study \label{sec:ablation}}

To further study the effect of \method, we present the ablation study on the \cona module and Focus-Consistent Constraint. All the experiments are conducted on C-GQA dataset with the same hyper-parameters.

\begin{table}[t]
\tablestyle{5pt}{1.0}
\setlength\tabcolsep{4pt}
\def\w{20pt} 
\scalebox{1}{
    \begin{tabular}{lc|cccccc}   
    Strategy& $\text{N}_\text{f}$&\textbf{A}&\textbf{O} & \textbf{S} & \textbf{U} & \textbf{AUC} & \textbf{HM}  \\
    \shline
    Standard &1& 12.5 & 30.8 & 30.9 & 14.6 & 3.4 &14.2\\
    Random &3&\textbf{13.6}&29.5&29.4&13.9&3.2&13.6\\
    Mean &3 &13.1&30.3&31.3&14.3&3.6&14.3\\
    \rowcolor[rgb]{.949, .949, .949} Predicted &3&13.3&\textbf{31.8}& \textbf{31.4}&\textbf{15.2}&\textbf{3.8}&\textbf{15.2} \\
    \end{tabular}
    }
\caption{\textbf{Analysis of the aggregation strategy} ($\text{N}_\text{f}$: \#~number of used feature, A: \#~attribute accuracy, O: \#~object accuracy). Compared to manually selected features and other aggregation strategies, our predictor yields the best~results.} 
\label{table:ab_agg_strategy}
\end{table}

\noindent\textbf{Is aggregation predictor actually predicting?} As the richness of inputs increases, the performance of the model is expected to improve. Therefore, it is insufficient to evaluate the competence of the predictor in the \cona module solely based on the improvement in performance. To demonstrate the effectiveness of the aggregation predictor, we compared our results with other strategies, i.e. randomly selecting the combination of features, only using the highest-level feature (noted with "standard") and computing the average of all features. As shown in \cref{table:ab_agg_strategy}, random selection yields inferior results, while "standard" and computing the mean are modest but fall short of fully exploiting the distinctive characteristics of the features. Our aggregation predictor, on the other hand, proves its effectiveness by outperforming the aforementioned strategy.

\noindent\textbf{What is the effect of Focus-Consistent Constraint?} To attest to the importance of spatial information and evaluate the effectiveness of the proposed constraint, we present the results of three ablations in \cref{table:ab_fcc}: one without Focus-Consistent Constraint, one with the attention pooling replaced by GAP and the other without the focused loss. The results show that the performance in the first row and second are similar, because the focused loss fails to take effect when used in conjunction with GAP. Besides, simply replacing GAP with attention pooling can lead to an improvement due to its ability to capture spatial information. In the end, attention pooling with the focused loss performs best, manifesting its advantage in constraining the three branches to focus on the consistent informative region and guiding the model to learn robust representations.

\begin{table}[!t]
\tablestyle{5pt}{1.0}
\setlength\tabcolsep{4pt}
\def\w{20pt} 
\scalebox{1}{
    \begin{tabular}{p{0.8cm}<{\centering}p{0.8cm}<{\centering}|cccccc}
    \textbf{AP}&$\textbf{L}_{\textbf{f}}$ & \textbf{A} & \textbf{O} & \textbf{S} & \textbf{U} & \textbf{AUC} &\textbf{HM}\\
    \shline
          &         &\textbf{14.2}  &30.0  &30.7  &13.6  &3.4  &14.1 \\
          & \cmark  &13.6  &30.4  &30.8  &13.6  &3.4  &14.0 \\
    \cmark &       &13.8  &30.5  &\textbf{31.4}  &14.1  &3.6  &14.6\\
    \rowcolor[rgb]{.949, .949, .949}
    \cmark & \cmark&13.3&\textbf{31.8}& \textbf{31.4}&\textbf{15.2}&\textbf{3.8}&\textbf{15.2} \\
    \end{tabular}
    }
  \caption{\textbf{Analysis of the effect of Focus-Consistent Constraint.} AP and $\text{L}_\text{f}$ stand for attention pooling and the focused loss. Applying attention pooling with the focused loss simultaneously leads to higher performance.}
  \label{table:ab_fcc}%
\end{table}%
\begin{figure}[t]
\begin{center}
\includegraphics[width=.7\textwidth]{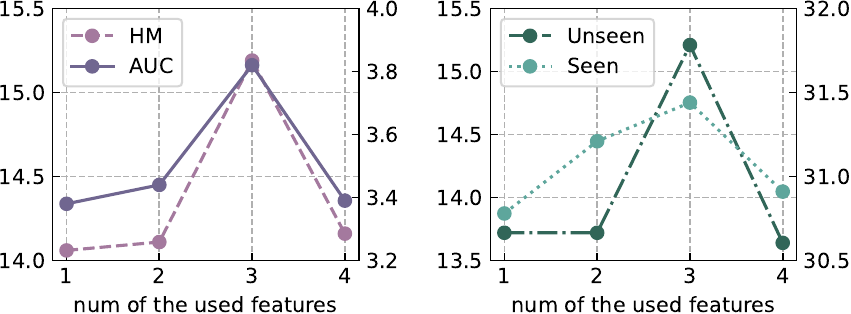}
\end{center}
\vspace{-3mm}
   \caption{\textbf{Analysis of the used features in ResNet-18.} The value of 4 indicates the use of all features, while 2 and 1 correspond to the use of the last two layers' features and the deepest layer's features, respectively. Our experiments show that incorporating the last three features from ResNet-18 yields the highest performance.}
\label{fig:ab_used_feats}
\end{figure}
\noindent\textbf{Why not use all the features in ResNet-18?} In the \cona module, we only allow the model to predict the weights of the last three levels of features, ignoring the feature from the first block of the backbone. This decision is based on the assumption that the first level feature is too coarse to convey sufficient information. As shown in \cref{fig:ab_used_feats}, this assumption is supported by our experiments, where integrating the first level feature leads to reduced performance at the cost of increased computational complexity. Beyond that, the model's performance declines as fewer features are used, indicating that all three features are valuable.

\begin{figure*}[!t]
\begin{center}
\includegraphics[width=1.0\textwidth]{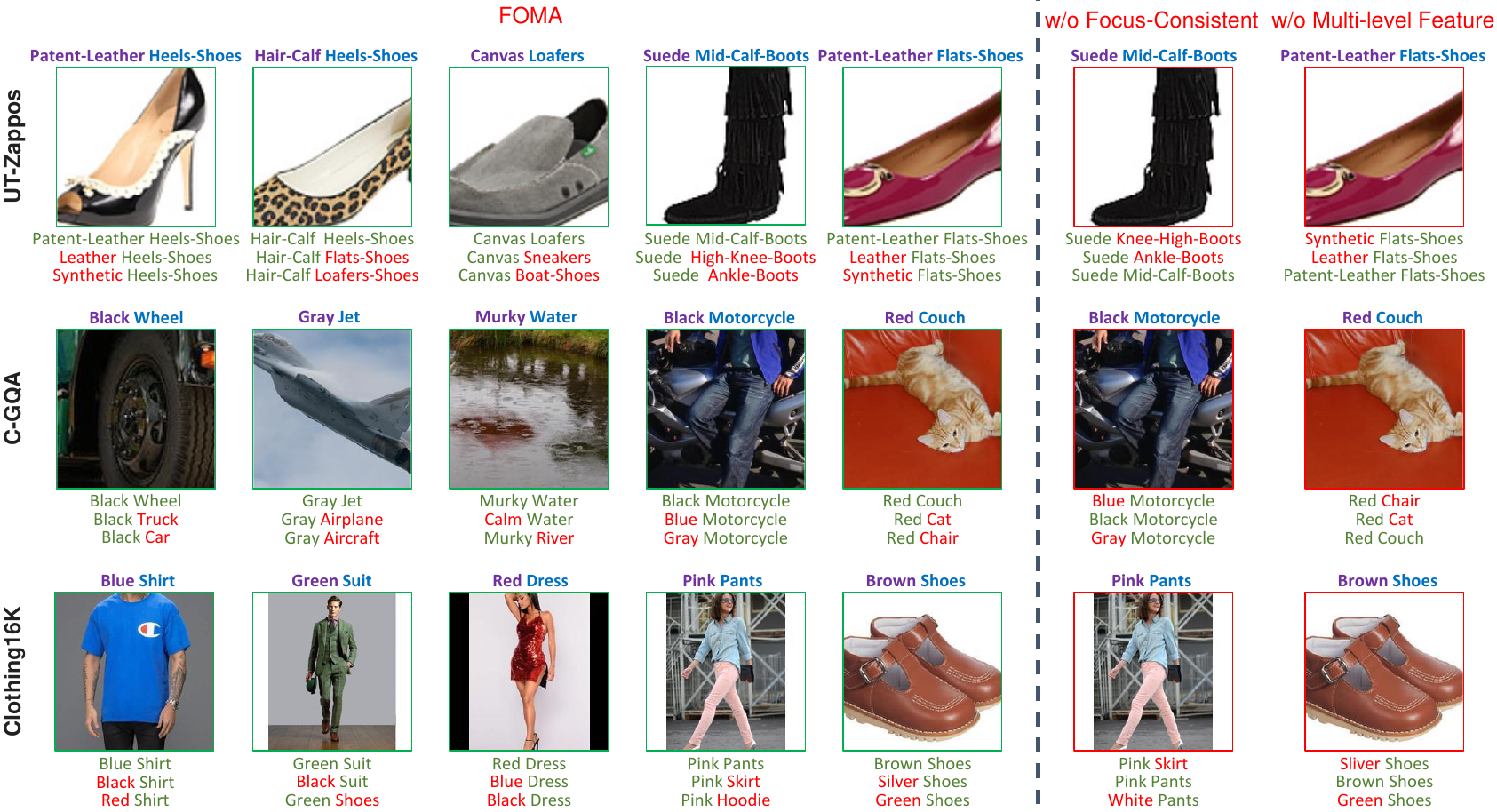}
\end{center}
\vspace{-3mm}
   \caption{\textbf{Qualitative Result.} We present the top-3 predictions from the three datasets. The first six columns display examples in which our model performed accurately, while the last two columns showcase errors produced by the two incomplete models. The correct and incorrect results are marked in green and red borders.}
\label{fig:qualitative}
\vspace{-3mm}
\end{figure*}


\subsection{Qualitative Results}

In order to evaluate the effectiveness of \method, we report the top-3 unseen compositional labels retrieved in \cref{fig:qualitative}.  On the left, we provide examples where the labels match the predictions made by our model. Careful readers should notice that one sample may also correspond to a couple of compositional labels as an object may be in different states or some of the words have synonymous meanings. For example, the third sample in the second row can be labeled as \textit{gray airplane} or \textit{gray aircraft}. This demonstrates that our model can generate multiple appropriate labels if they exist.

On the right, examples are provided to illustrate the incorrect predictions made by two types of incomplete models. The last two columns present the results obtained from the model without Focus-Consistent Constraint and the model without \cona module, i.e. employing the feature from the final layer of ResNet-18. As we can see, incomplete models would make some mistakes when faced with indistinguishable samples. Take \textit{Red Couch} in the second row for example. The model without multi-level features overlooked the texture of the leather and misidentified the sample as a chair instead of a couch. In contrast, the complete model successfully matches the label, thanks to the clues from lower-level features. All of the incorrect examples highlight the significance of Focus-Consistent Constraint and \cona module.

\begin{figure}[!t]    
  \centering           
  \subfloat[]
  {
      \label{fig:quan_weight_and_visual_feats_a}\includegraphics[width=0.5\textwidth]{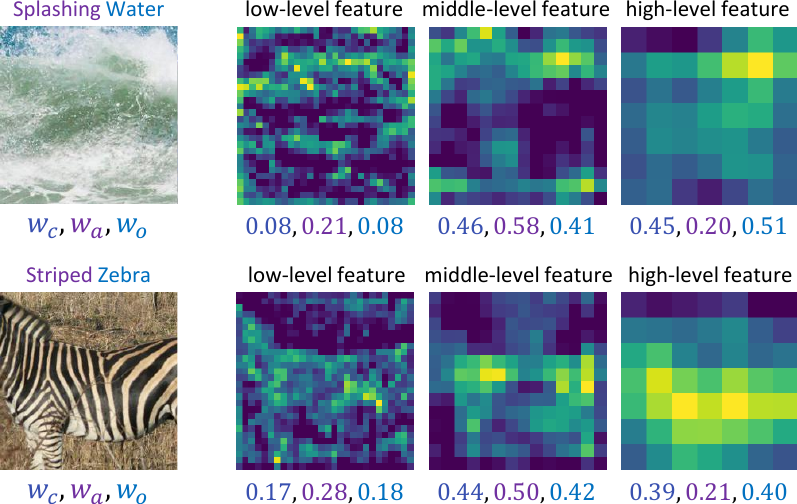}
  }
  \subfloat[]
  {
      \label{fig:quan_weight_and_visual_feats_b}\includegraphics[width=0.47\textwidth]{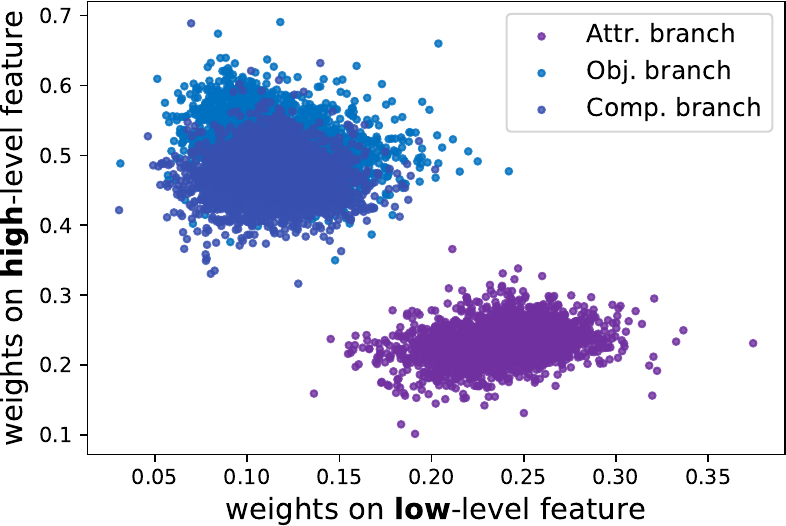}
  }
\caption{\textbf{(a) Two examples of aggregation weights and feature maps.} The three columns on the right are the visualization of low-level feature, middle-level feature and high-level feature. Numbers under the feature maps correspond to the weights assigned to each branch. Lower-level features and higher-level features exhibit a stronger correlation with  the detailed regions and the global patterns. \textbf{(b) Statistical analysis of the aggregation weights.} The attribute branch incline to acquire higher weights on low-level feature, whereas the weights from the object branch exhibit contrasting characteristics. The composition branch weights are situated between the two. }
\vspace{-5mm}
\label{fig:quan_weight_and_visual_feats} 
\end{figure}

In order to better illustrate the functionality of the aggregation predictor, we present the output of this predictor through two examples, as depicted in \cref{fig:quan_weight_and_visual_feats_a}. It is obvious that the predictor assigns various weights to each branch depending on the input. Typically, the object branch places greater emphasis on higher-level features, while the attribute branch gains larger weight on lower-level features. Moreover, the weights for the composition branch are approximately equal to the mean of the other two weights. For example, \textit{splashing water} receives the higher weights on low-level feature and middle-level feature in the attribute branch. We assume that this is because \textit{splashing} is more related to the detailed regions than the global patterns. On the other hand, the model requires a global perspective to recognize \textit{water}, which explains why middle-level feature and high-level feature acquire higher weights in the object branch.

Apart from that, we analyze the statistical results of aggregation weights. As is shown in \cref{fig:quan_weight_and_visual_feats_b}, we plot a scatter diagram of the weights obtained from the test set of C-GQA. Each dot in the diagram represents an image. It is apparent that the object branch gives priority to higher-level features, whereas the attribute branch places more emphasis on lower-level features. The composition branch weights balance the importance of the attribute and the object branches by attaining relatively placid weights. This phenomenon aligns with our assumption without any deliberate planning.

\section{Hyper-Parameter Analysis}\label{sec:hyper-para}
\begin{figure}[t]
  \centering
  \begin{minipage}[t]{1\linewidth}
      \centering
      \includegraphics[width=5in]{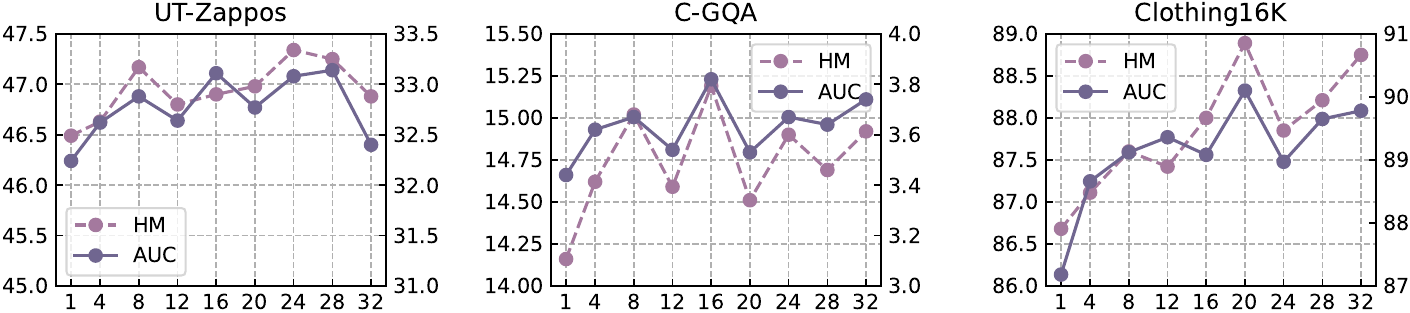}
      \caption{Hyper-parameter analysis of the temperature $\tau$.}
      \label{fig:hyper_a}
  \end{minipage}
  \begin{minipage}[t]{1\linewidth}
      \centering
      \includegraphics[width=5in]{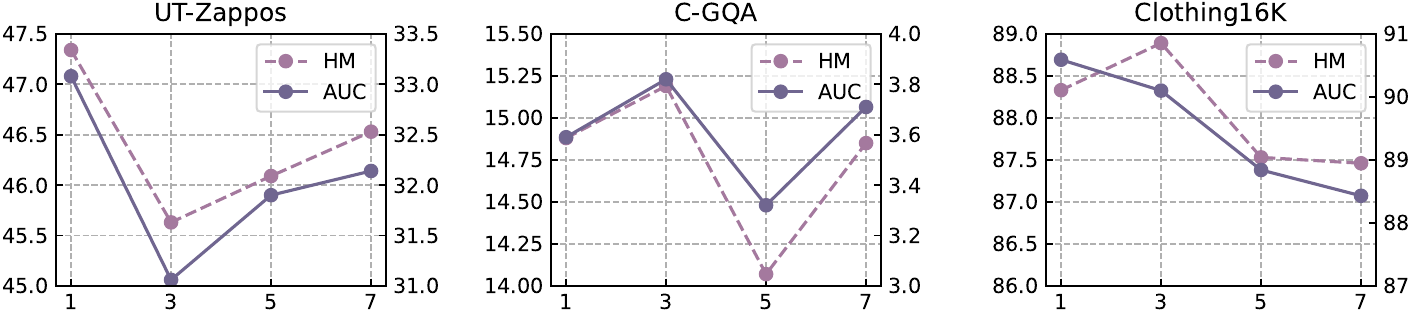}
      \caption{Hyper-parameter analysis of  the loss coefficient $\alpha$.}
      \label{fig:hyper_b}
  \end{minipage}
\vspace{-3mm}

\end{figure}

In this section, we present HM and AUC for different values of the loss coefficient $\alpha$ and the temperature $\tau$, as illustrated in \cref{fig:hyper_a} and \cref{fig:hyper_b}. In the case of the loss coefficient $\alpha$, we observe that the results of the model exhibit limited variations across a range of values from 1 to 7. What's more, we can draw a similar conclusion regarding the effect of the temperature $\tau$. As $\tau$ increases from 1 to 32, given that the aggregation predictor is capable of adjusting its parameters through backpropagation, our model attains a relatively consistent performance. These experiments suggest that our model is robust to changes in these hyper-parameters. In the end, We determine the values of $\alpha$ and $\tau$ to be 1, 3, 3 and 24, 16, 20 for UT-Zappos, C-GQA, and Clothing16k datasets.

\section{Conclusion}

In this work, we propose a novel Focus-Consistent Multi-Level Aggregation (\method) method to address the compositional zero-shot learning task. 
Our methods containing three branches can make full use of multi-level features and spatial messages. 
To begin with, Multi-Level Feature Aggregation (\cona) module generates instance-specific features for particular recognition. 
Besides, Focus-Consistent Constraint on the three branches to correct the process if one branch is influenced by irrelevant regions. 
The experiments demonstrate that \method can learn robust primitives during training and generalize to unseen compositions. We show the advantage of \method by outperforming state-of-the-art methods on three widely used benchmark datasets.



\bibliographystyle{splncs04}
\bibliography{main}

\end{document}